\documentclass[11pt]{article}
\PassOptionsToPackage{hyphens}{url}
\usepackage[preprint]{acl}

\usepackage{times}
\usepackage{latexsym}
\usepackage[T1]{fontenc}
\usepackage[utf8]{inputenc}
\usepackage{microtype}
\usepackage{inconsolata}
\usepackage{graphicx}
\usepackage{amsmath,amssymb}
\usepackage{booktabs}
\usepackage{multirow}
\usepackage{makecell}
\usepackage{xcolor}
\usepackage{enumitem}
\usepackage{listings}
\usepackage{tikz}
\usetikzlibrary{positioning,arrows.meta,fit,backgrounds,calc}
\pgfdeclarelayer{bottom}\pgfsetlayers{bottom,background,main}
\usepackage{pgfplots}
\pgfplotsset{compat=1.18}
\usepackage{cleveref}

\makeatletter
\renewcommand\@makefntext[1]{\noindent\@makefnmark\raggedright#1}
\makeatother
\newcommand{\solfull}{GPT-6.1-Sol}
\newcommand{\sysname}{Delegent}                           % name of our harness
\newcommand{\qwen}{Qwen3.8-27B-FP8}
\newcommand{\mini}{MiniCPM5-2B}
\newcommand{\solsh}{Sol}
\newcommand{\opussh}{Opus}
\newcommand{\qwensh}{Qwen}
\newcommand{\minish}{MiniCPM}
\newcommand{\opus}{Claude Opus 5.5}

\newsavebox{\promptbin}
\newenvironment{promptbox}{\par\medskip\noindent\setlength{\fboxsep}{4pt}\begin{lrbox}{\promptbin}%
  \begin{minipage}{\dimexpr\columnwidth-2\fboxsep-2\fboxrule\relax}\scriptsize\ttfamily\raggedright
  \microtypesetup{protrusion=false}%
  \setlength{\parskip}{3pt}\setlength{\parindent}{0pt}}%
  {\end{minipage}\end{lrbox}\fcolorbox{black!30}{black!4}{\usebox{\promptbin}}\par\medskip}
\newcommand{\pitem}[1]{\par\hangindent=1.2em\hangafter=1 -\ #1}

\title{Delegent: A Multi-Agent Harness\\That Reduces Cost without Sacrificing Performance}

\author{Lingwei Gu \quad Jimmy Lin\\[1ex]
  David R. Cheriton School of Computer Science\\
  University of Waterloo}

\begin{document}
\maketitle

% ===========================================================================
\begin{abstract}
Agents built on frontier language models are expensive. At every step, the frontier model re-reads its entire context and
executes every subtask regardless of its difficulty. This paper explores the simple thesis that much of this
cost can be removed by delegating easy subtasks to a small language model, or SLM, that runs on hardware one already owns.
To test it, we build \sysname, a simple and effective agent harness built on Claude Code, in which the frontier model stays in charge of
every task but hands focused subtasks, such as investigating code, running tests, and searching a corpus, to SLM
subagents. Existing approaches such as cascades and model routers usually choose which model executes the entire task, whereas in \sysname, the frontier model always leads the execution and delegates easy subtasks to the SLM. No router or model is trained, so either model can be swapped freely. With the help of
a 27B FP8 SLM, \sysname\ reduces the frontier cost of long-horizon coding by 67\% and of agentic search by 59\%, with no loss
in task performance, and even a 2B SLM that fits on a laptop reduces the coding cost by 66\%. The savings grow with the legwork that the frontier model would otherwise do itself, and a short
instruction, which leaves the main coding tasks to the frontier model, accounts for most of the saving and keeps the 2B SLM
from losing tasks. Code: \url{https://github.com/castorini/delegent}.
\end{abstract}

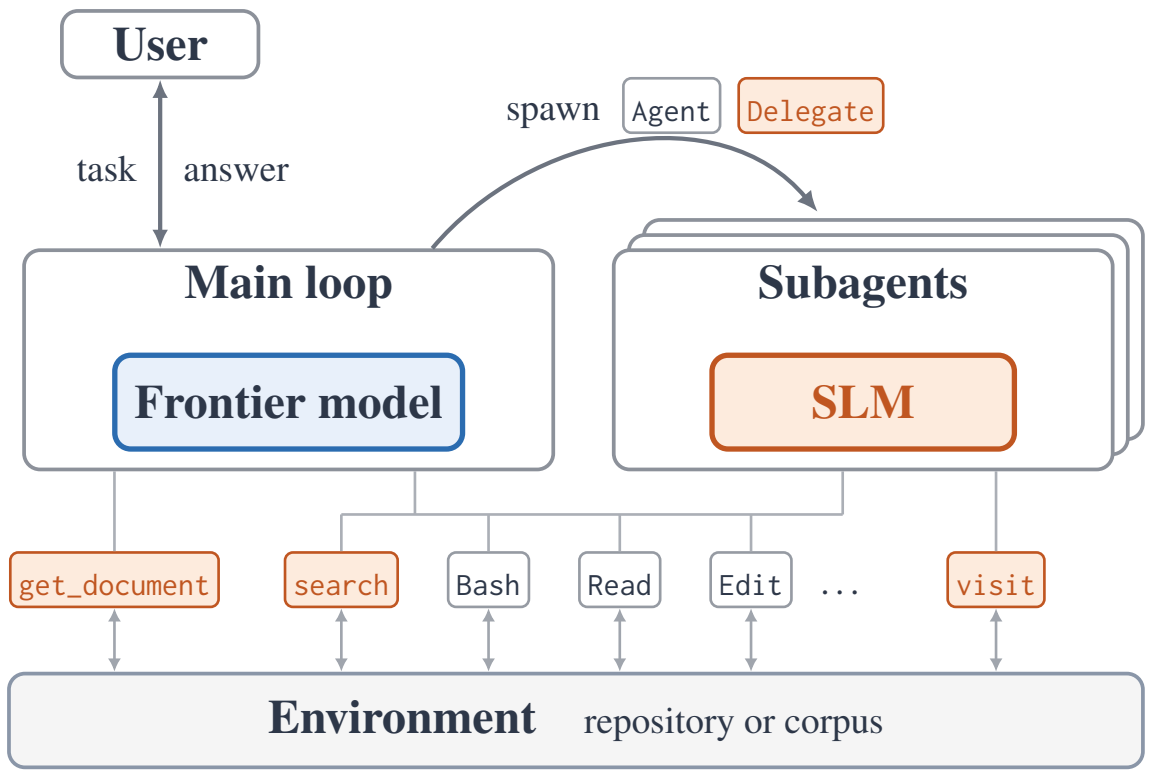
\begin{figure}[t]
\centering
\definecolor{fblue}{HTML}{2B6CB0}\definecolor{fbluebg}{HTML}{E8F0FA}
\definecolor{horange}{HTML}{C05621}\definecolor{horangebg}{HTML}{FDEBDD}
\definecolor{ink}{HTML}{2D3748}\definecolor{soft}{HTML}{8A96A8}
\begin{tikzpicture}[font=\footnotesize, every node/.style={text=ink},
  box/.style={draw, rounded corners=3pt, line width=0.7pt, align=center, inner xsep=3pt, inner ysep=2pt},
  agent/.style={box, draw=ink!55, fill=white, minimum height=1.45cm, anchor=north},
  lab/.style={font=\scriptsize, align=center, inner sep=1pt, text height=1.6ex, text depth=0.3ex},
  tool/.style={draw=ink!50, fill=white, line width=0.5pt, rounded corners=1.5pt, font=\fontsize{6}{7}\selectfont\ttfamily,
    inner xsep=1.6pt, inner ysep=0pt, minimum height=0.36cm, text height=1.5ex, text depth=0.35ex, anchor=south west},
  new/.style={tool, draw=horange, fill=horangebg, text=horange},
  dots/.style={lab, anchor=south west, minimum height=0.36cm},
  call/.style={{Latex[length=3.5pt,width=3pt]}-{Latex[length=3.5pt,width=3pt]}, line width=0.8pt, line cap=round},
  use/.style={line width=0.5pt, draw=ink!40},
  act/.style={{Latex[length=2.5pt,width=2.5pt]}-{Latex[length=2.5pt,width=2.5pt]}, line width=0.5pt, draw=ink!45}]
% main loop
\node[agent, minimum width=3.5cm] (M) at (1.85,0) {};
\node[anchor=north, font=\footnotesize\bfseries, inner sep=3pt] at (M.north) {Main loop};
\node[box, draw=fblue, fill=fbluebg, line width=1pt, minimum width=2.3cm, minimum height=0.62cm, anchor=south]
  (F) at ($(M.south)+(0,0.13)$) {\textbf{Frontier model}};
% subagents: a stack
\foreach \k in {2,1} {\node[agent, minimum width=3.3cm] at ($(5.65,0)+(0.1*\k,0.1*\k)$) {};}
\node[agent, minimum width=3.3cm] (S) at (5.65,0) {};
\node[anchor=north, font=\footnotesize\bfseries, inner sep=3pt] at (S.north) {Subagents};
\node[box, draw=horange, fill=horangebg, line width=1pt, minimum width=2.0cm, minimum height=0.62cm, anchor=south]
  (L) at ($(S.south)+(0,0.13)$) {\textcolor{horange}{\textbf{SLM}}};
% spawn, through the Agent tool (coding) or the Delegate tool (search)
\draw[-{Latex[length=4pt,width=3.5pt]}, line width=0.8pt, draw=ink!70] ($(M.north)+(0.95,0)$) to[out=50, in=130]
  coordinate[pos=0.5] (apex) ($(S.north)+(-0.3,0.24)$);
\node[lab, anchor=south east] (sp) at ($(apex)+(-0.12,0.06)$) {spawn};
\node[tool, anchor=south west] (ag) at ($(apex)+(-0.02,0.06)$) {Agent};
\node[new, right=3pt of ag] {Delegate};
% tools, each used by the loop above it and acting on the environment below
\coordinate (R) at (0,-2.38);
\node[new] (g1) at (0.0,0 |- R) {get\_document};
\node[new, right=12pt of g1] (s1) {search};
\node[tool, right=9pt of s1] (b1) {Bash};
\node[tool, right=9pt of b1] (r1) {Read};
\node[tool, right=9pt of r1] (e1) {Edit};
\node[dots] (d1) at ($(e1.south east)+(4pt,0)$) {\ldots};
\node[new] (v1) at (6.2,0 |- R) {visit};
\coordinate (bus) at (0,-1.76);
\node[box, draw=soft, fill=gray!8, minimum width=7.5cm, minimum height=0.62cm, anchor=north] (E) at (3.75,-2.8)
  {\textbf{Environment}\quad{\scriptsize repository or corpus}};
% shared tools hang from a bus that both loops reach; loop-specific tools connect to their own loop only
\draw[use] (s1.north |- bus) -- (d1.north |- bus);
\coordinate (mx) at ($(s1.north)!0.5!(b1.north)$);
\draw[use] (mx |- bus) -- (mx |- M.south);
\draw[use] (d1.north |- bus) -- (d1.north |- S.south);
\foreach \t in {s1,b1,r1,e1} {\draw[use] (\t.north) -- (\t.north |- bus);}
\draw[use] (g1.north) -- (g1.north |- M.south);
\draw[use] (v1.north) -- (v1.north |- S.south);
\foreach \t in {g1,s1,b1,r1,e1,v1} {\draw[act] (\t.south) -- (\t.south |- E.north);}
% user
\node[box, draw=ink!55, fill=white, minimum width=1.3cm, minimum height=0.44cm] (U) at (1.0,1.35) {\textbf{User}};
\draw[call, draw=ink!70] (U.south) -- node[lab, left=3pt] {task} node[lab, right=3pt] {answer} (U.south |- M.north);
\end{tikzpicture}
\caption{The \sysname\ harness. It starts from the stock Claude Code harness, and everything in orange is what we
add. The user talks to the main loop, where the frontier model leads the task and spawns subagents. The loops share
most tools, and each adds a few.}
\label{fig:overview}
\end{figure}

% ===========================================================================
\section{Introduction}
\label{sec:intro}

Agents are simply models using tools in loops. The harness invokes the model, executes its tool calls, returns the
results, and invokes the model again, until the task is done. The loop is simple but costly, because at every step the model
re-reads its entire context, which grows with every tool call. A frontier coding agent that takes a hundred steps to
resolve an issue thus pays for a hundred ever longer prompts. Long-horizon agents are therefore expensive for organizations that pay per
token, and individuals on subscription plans exhaust their quota and have to wait for the next reset.

Yet much of what a frontier model does at each step is easy legwork, such as locating the relevant code, reproducing a bug, running
tests, issuing searches, and reading documents. Little of this work seems to require a frontier model, yet existing agents use the
frontier model for all of it, even in subagents, billing every step at frontier prices. A small language model, or SLM, costs far
less, and nothing extra on a GPU that one already owns. An SLM on its own falls well short of a frontier model on these tasks and cannot
replace one, but it does not need to compete with the frontier if the frontier model stays in charge and delegates only easy subtasks. This leads to a simple thesis:
\begin{quote}
\interlinepenalty=10000 % keep the thesis in one column
\emph{A small language model running on local hardware can take over the easy subtasks of a frontier model, cutting its
cost substantially without hurting task performance, and without any training.}
\end{quote}

\noindent To test this thesis, we build \sysname\ (\Cref{fig:overview}), an agent harness that runs a frontier model and a local SLM as a team. For
coding, \sysname\ is derived from Claude Code.\footnote{\url{https://code.claude.com/docs}} It keeps the agent loop and subagents of
Claude Code, but routes each model call by role, sending the calls of the main loop to the frontier model and the calls
of its subagents to the SLM. For agentic search, it adds a \texttt{Delegate} tool whose subagents search the corpus and
return reports with evidence. The frontier model stays in charge of every task. A short instruction tells it what to delegate, the harness runs
the delegated subtasks on the SLM and checks the evidence in their reports, and the frontier model alone produces the
final answer. Since nothing is trained, either model is a drop-in component.

As the headline result, we find that the thesis holds on both long-horizon coding and agentic search. With \solfull\ at
high reasoning effort as the frontier model and \qwen\ as the SLM, the frontier model's cost on DeepSWE falls to a third,
with no loss in resolved tasks. On 694 search questions, its cost falls to 41\%, and it answers exactly as many
questions correctly as on its own. Even a 2B model, \mini, cuts frontier cost on DeepSWE to a third, and with \qwen\ paid for at API prices, the
total cost of a DeepSWE task is still only 38\%. Delegation adds
some latency, which is bounded by the local hardware that serves the SLM.

Contrary to the dominant approach to combining cheap and expensive models, we do not let the SLM answer first and then
decide whether to trust it~\citep{chen2023frugalgpt,madaan2023automix,yue2024cascades}, nor do we train a router that
picks one model per query~\citep{ding2024hybridllm,ong2024routellm}. In both cases, a model is given the entire task and must solve it by itself, which a 2B SLM is rarely capable of. In \sysname, the frontier model always produces the
final answer, and the SLM only does the legwork. No output of the SLM is accepted without the frontier model reading it
first. Our contribution is thus a piece of harness engineering: the coordination of two models of very
different cost.

To better circumscribe our claims, let us be more precise:
\begin{itemize}[leftmargin=*,itemsep=1pt,topsep=2pt,parsep=0pt,partopsep=0pt]
  \item By \emph{cost}, we mean what the frontier model is charged, with prompt caching taken into account. The SLM
        runs on a GPU under the user's control, and for coding, we also price \qwen\ at API prices.
  \item By \emph{local}, we mean an SLM served on a single GPU under the user's control. The 27B FP8 model needs a
        workstation GPU, and the 2B model fits on a laptop.
  \item By \emph{training-free}, we mean that no model is trained or fine-tuned, so the SLM and the frontier
        model can each be swapped freely.
\end{itemize}

\noindent This paper makes three contributions. First, we build \sysname, an open-source, training-free harness in which a frontier
model delegates work to a local SLM, which we describe in \Cref{sec:method}. Second, we evaluate it on three coding and
five search datasets with two SLMs of very different sizes, and show in \Cref{sec:results} that frontier cost falls by
more than half in eight of nine evaluations while task performance holds. Third, we analyze in \Cref{sec:analysis} where the savings come from and how they depend on our instruction and
the frontier model.

% ===========================================================================
\section{The \sysname\ Harness}
\label{sec:method}

\sysname\ is built on Claude Code and coordinates two models of very different cost, as \Cref{fig:overview} illustrates. The \emph{frontier model}
leads every task. It plans, decides what to delegate and whether to believe what comes back, and writes the final patch
or answer. \emph{SLM subagents} on a local GPU do the legwork, each in its own tool loop and with a fresh context.
\sysname\ lets the frontier model judge the task type, coding or search, and supplies the matching tool set.

\paragraph{Routing model calls by role.}
In Claude Code, the main loop calls a model at every step, and so does each subagent that it starts with the built-in Agent
tool. Minor jobs go to a small auxiliary model. For each task, \sysname\ places a relay between Claude Code and the models. The relay speaks
the Anthropic Messages API, so Claude Code runs unmodified, and it routes each call according to the model that the call
names. Calls from the main loop go to the frontier model, and calls from the subagents and the auxiliary model go to
the SLM. Every frontier call within a task carries the same prompt-cache key, so consecutive steps hit the provider's
cache. SLM calls go to vLLM~\citep{kwon2023vllm} through a proxy that pins each subagent to one GPU replica, so
its growing prefix stays cached.

\paragraph{Running subtasks.}
For coding, the frontier model delegates a subtask through its Agent tool and writes a brief, the instructions for the
Claude Code subagent. The subagent works in the same container with Claude Code's default tools. For search, \sysname\ adds a \texttt{Delegate} tool for delegating one focused search
subtask, such as finding or checking a fact. Instead of designing the search subagent from scratch, we borrow from
AREX~\citep{baai2026arex}, a family of deep research agents. Its smallest member, AREX-Turbo, is built around a 4B
model, so its prompts and tools are already written for a small model to follow. An AREX-Turbo agent works in a loop
with four tools: it searches the corpus, visits the documents it finds, compresses its own history into notes when the
history grows long, and finishes with an answer, the evidence that supports it, and a confidence. Our SLM subagent sees only the subtask
and the question, and its final answer, evidence, and confidence form the report that returns to the frontier model. In both coding and search, subtasks from one turn run concurrently.

\paragraph{Returning reports.}
\looseness=-1 The frontier model's context receives only the final report of a subagent and none of its tool outputs. For search, the harness
also looks up each evidence line in the cited document and notes whether it was found verbatim.

\paragraph{Steering the frontier model.}
A short instruction (\Cref{app:prompts}) explains the economics to the frontier model. Since every step re-reads the entire context, the
cheapest plan has the fewest steps of the frontier model's own, so the instruction asks it to delegate tool-heavy work
and to keep its own tool use to a few targeted checks. It covers both kinds of delegation, Claude Code's subagents for
coding and the \texttt{Delegate} tool for search.

% ===========================================================================
\section{Experimental Setup}
\label{sec:setup}

\paragraph{Tasks.}
For coding, we use random samples of SWE-bench
Verified~\citep{jimenez2024swebench},\footnote{\url{https://openai.com/index/introducing-swe-bench-verified/}} SWE-bench Pro~\citep{deng2025swebenchpro}, and DeepSWE
v1.1~\citep{deepswe2025}, drawn as described in \Cref{app:sampling}.  For search, we use 200 BrowseComp-Plus questions~\citep{chen2025browsecompplus},
randomly sampled from those not used to draft the instruction, and four datasets from CMASS~\citep{cmass2026}, an agentic search benchmark over
ClimbMix~\citep{diao2025climb}, an open web corpus of 400B tokens that agents can access only through BM25
search~\citep{robertson2009bm25,lin2021pyserini}.  The CMASS datasets are BrowseComp-Plus, Total Recall
QA~\citep{rafiee2026trqa}, HotpotQA~\citep{yang2018hotpotqa}, and DeepSearchQA~\citep{gupta2026deepsearchqa}, which we
abbreviate as TRQA and DSQA. DSQA questions have several correct answers, and CMASS evaluates them in two modes. In \emph{find one}, any correct answer counts, while \emph{find all}
needs the exact set.

\paragraph{Models.}
All our runs use the \sysname\ harness, built on Claude Code, and the frontier model it calls can change
freely. Our main experiments call \mbox{\solfull}, accessed through the OpenAI API, and on DeepSWE we run ablations
with \mbox{\opus} at xhigh effort. We run \solsh\ at high reasoning effort, the same with and without delegation, to keep our
experiments affordable. At this effort, it resolves just over half of the DeepSWE tasks alone. We pair it with two SLMs of very different sizes, both easy to run on a single
device. \qwen\footnote{\url{https://qwen.ai/blog?id=qwen3.8}} fits on one workstation GPU, and \mini~\citep{hu2024minicpm,minicpm2025}
is small enough for a laptop. We
serve both with vLLM, one GPU per replica, and test \mini\ on coding only.

\paragraph{Baseline.}
We compare against the frontier model alone on the same tasks, using the same harness without the delegation tool and instruction.
For coding, this is Claude Code with its default tools and subagents, where we send every call to the frontier model, including the calls
that Claude Code sends to Haiku by default. As a sanity check, \opus\ alone resolves
68\% of our DeepSWE tasks in this setup, in line with the 68.4\% that Artificial Analysis reports for \opus\ in Claude Code.\footnote{\url{https://x.com/ArtificialAnlys/status/2102932119995756613}, for \opus\ at max effort on DeepSWE v1.1.}

\paragraph{Metrics.}
A coding task is resolved if the final patch passes the task's hidden tests. A search answer is graded against the gold
answer by \solfull\ at low effort with the official BrowseComp-Plus grading prompt~\citep{chen2025browsecompplus}, and
the same judge grades every system. All coding runs, including those of the frontier model alone, use Claude Code with its default tools except web access, a
context of up to 1M tokens for the frontier model and 131k tokens for the SLM, and the same limits on turns and time,
which \Cref{app:settings} lists. Latency is wall-clock time per task.

\paragraph{Cost.}
We count frontier cost at the API prices of the frontier model,\footnote{\url{https://developers.openai.com/api/docs/pricing}, accessed October 2026.} with prompt caching for
\mbox{every system}:
\begin{equation*}
  \text{cost} = \$2\,T_{\text{in}}^{\text{uncached}} + \$0.10\,T_{\text{in}}^{\text{cached}} + \$10\,T_{\text{out}},
\end{equation*}
where $T$ are token counts in millions and output includes reasoning tokens. We report cost per task and relative to the
frontier model alone. For coding, we also price \qwen\ at the API rates of Chutes, a provider that serves the same FP8
model, with prompt caching as well. Input costs \$0.039 per million tokens, a tenth of that when it repeats the
subagent's previous request, and output costs \$1.25 per million tokens.

% ===========================================================================
\section{Results}
\label{sec:results}

\begin{table*}[t]
\centering
\small
\setlength{\tabcolsep}{2.5pt}
\begin{tabular}{@{}llcccccc@{}}
\toprule
Task set & System & Solved & \makecell{Frontier cost\\(\$ per task)} & \makecell{Relative\\cost} & \makecell{Cost with SLM\\at API price} & \makecell{Frontier calls\\per task} & \makecell{SLM calls\\per task} \\
\midrule
\multirow{3}{*}{Verified (50)}
 & Frontier model alone & 46 & 0.244 & 100\% & 0.244 (100\%) & \phantom{0}32.4 & -- \\
 & + \qwen & 46 & 0.108 & \phantom{0}44\% & 0.139 (57\%) & \phantom{0}12.7 & \phantom{0}32.7 \\
 & + \mini & 47 & 0.129 & \phantom{0}53\% & -- & \phantom{0}15.2 & \phantom{0}79.0 \\
\midrule
\multirow{3}{*}{Pro (75)}
 & Frontier model alone & 63 & 0.617 & 100\% & 0.617 (100\%) & \phantom{0}61.2 & -- \\
 & + \qwen & 63 & 0.227 & \phantom{0}37\% & 0.266 (43\%) & \phantom{0}19.9 & \phantom{0}42.7 \\
 & + \mini & 62 & 0.274 & \phantom{0}44\% & -- & \phantom{0}23.1 & \phantom{0}65.5 \\
\midrule
\multirow{3}{*}{DeepSWE (50)}
 & Frontier model alone & 29 & 2.110 & 100\% & 2.110 (100\%) & 176.1 & -- \\
 & + \qwen & 29 & 0.704 & \phantom{0}33\% & 0.806 (38\%) & \phantom{0}46.6 & \phantom{0}82.0 \\
 & + \mini & 30 & 0.712 & \phantom{0}34\% & -- & \phantom{0}48.1 & 106.1 \\
\bottomrule
\end{tabular}
\caption{Coding results with \solfull\ as the frontier model. Relative cost is relative to the frontier model alone;
cost with SLM adds the SLM at the API price of \qwen\ (\Cref{sec:setup}); no API serves \mini.}
\label{tab:coding}
\end{table*}

\begin{table*}[t]
\centering
\small
\begin{tabular}{@{}llccccc@{}}
\toprule
Task set & System & Solved & \makecell{Frontier cost\\(\$ per task)} & \makecell{Relative\\cost} & \makecell{Frontier calls\\per task} & \makecell{SLM calls\\per task} \\
\midrule
\multirow{2}{*}{BrowseComp-Plus$^\dagger$ (200)}
 & Frontier model alone & 193 & 0.114 & 100\% & \phantom{0}8.8 & -- \\
 & + \qwen & 192 & 0.071 & \phantom{0}63\% & \phantom{0}5.7 & 305.2 \\
\midrule
\multirow{2}{*}{BrowseComp-Plus (80)}
 & Frontier model alone & 72 & 0.211 & 100\% & 13.9 & -- \\
 & + \qwen & 71 & 0.066 & \phantom{0}31\% & \phantom{0}6.2 & 495.7 \\
\midrule
\multirow{2}{*}{TRQA (68)}
 & Frontier model alone & 64 & 0.048 & 100\% & \phantom{0}6.2 & -- \\
 & + \qwen & 67 & 0.015 & \phantom{0}31\% & \phantom{0}3.1 & \phantom{0}69.1 \\
\midrule
\multirow{2}{*}{HotpotQA (204)}
 & Frontier model alone & 198 & 0.033 & 100\% & \phantom{0}5.4 & -- \\
 & + \qwen & 198 & 0.008 & \phantom{0}25\% & \phantom{0}2.5 & \phantom{0}29.3 \\
\midrule
\multirow{2}{*}{DSQA, find one (71)}
 & Frontier model alone & 70 & 0.067 & 100\% & \phantom{0}7.7 & -- \\
 & + \qwen & 70 & 0.020 & \phantom{0}29\% & \phantom{0}3.6 & 106.2 \\
\midrule
\multirow{2}{*}{DSQA, find all (71)}
 & Frontier model alone & 66 & 0.121 & 100\% & 10.3 & -- \\
 & + \qwen & 65 & 0.032 & \phantom{0}26\% & \phantom{0}4.3 & 223.9 \\
\midrule
\multirow{2}{*}{All (694)}
 & Frontier model alone & 663 & 0.091 & 100\% & \phantom{0}8.2 & -- \\
 & + \qwen & 663 & 0.037 & \phantom{0}41\% & \phantom{0}4.2 & 194.2 \\
\bottomrule
\end{tabular}
\caption{Search results, with frontier cost per question. $^\dagger$Over its own corpus; all other sets over ClimbMix.}
\label{tab:search}
\end{table*}

\paragraph{Coding.}
The thesis holds on all three coding sets (\Cref{tab:coding}). With \qwen\ as its subagent, the
frontier model resolves no fewer tasks than on its own, at between a third and a half of the cost. On DeepSWE, the frontier model's cost falls
from \$2.11 to \$0.70 per task. The results with \mini\ are much the same: task performance holds, and on DeepSWE the 2B
model saves as much as the 27B FP8 model.

Paying for the SLM changes little on long tasks. At API prices, \qwen\ adds about 10 cents to a DeepSWE task, so its total cost stays at 38\%, and even on the short tasks of SWE-bench Verified, the total is 57\%.

\paragraph{Search.}
Across all 694 search questions (\Cref{tab:search}), the frontier model answers 663 correctly with \qwen\ as its subagent, exactly as many as on its own, at
41\% of the cost. On no set do the two systems differ by more than three questions. The savings, however, vary
considerably. On ClimbMix, frontier cost falls to between a quarter and a third, but on BrowseComp-Plus
over its own corpus, it falls only to 63\%.

The difference lies in how much the frontier model has to read on its own. On BrowseComp-Plus, the corpus is a set of
files, and a command such as \texttt{rg} returns only the lines that match. On ClimbMix, the only access is BM25
search, which returns whole documents that the model then reads a page at a time. Answering the same questions thus
takes more than twice as much reading on ClimbMix, and delegation moves most of it to the SLM. Since a web agent also
finds pages through a search engine and opens them, we consider the ClimbMix savings more representative.

\paragraph{Prompt caching.}
Every system, including the frontier model alone, uses the provider's prompt cache with a stable key per task, as any
practical deployment would, so all savings are measured against a cached baseline. Nearly all frontier input on coding is cached, with or without delegation.

\paragraph{Latency.}
Delegation adds latency that depends mainly on the local hardware serving the SLM. With \qwen, the median DeepSWE task takes 38 minutes instead of 25,
and a short SWE-bench Verified task about twice as long.

% ===========================================================================
\section{Analysis}
\label{sec:analysis}

The results leave two questions open, namely where the savings come from and why task performance holds. We answer
them with statistics computed from the transcripts of our runs. Unless noted otherwise, we use DeepSWE, the hardest of
our task sets. Its tasks are long, and the frontier model alone resolves just over half of them, so DeepSWE leaves the
most room for savings and would also reveal any loss in task performance.

\begin{figure}[t]
\centering
\begin{tikzpicture}
\begin{axis}[
  width=\columnwidth, height=3.1cm, ybar, bar width=14pt, ymin=0, ymax=0.92,
  symbolic x coords={<25,25--50,50--100,100--200,>200}, xtick=data,
  xlabel={Frontier model's calls on the task when working alone},
  ylabel={Relative cost}, label style={font=\small}, tick label style={font=\scriptsize},
  nodes near coords, nodes near coords style={font=\scriptsize}, every node near coord/.append style={/pgf/number format/fixed, /pgf/number format/precision=2},
  enlarge x limits=0.15, grid=major, grid style={gray!20}]
\addplot[fill=blue!35, draw=blue!60] coordinates {(<25,0.62) (25--50,0.47) (50--100,0.38) (100--200,0.36) (>200,0.26)};
\end{axis}
\end{tikzpicture}
\caption{Median relative frontier cost per coding task, by how many calls the frontier model makes on the task when
working alone. Pooled over 350 task pairs from the three coding sets and both SLMs.}
\label{fig:legwork}
\end{figure}
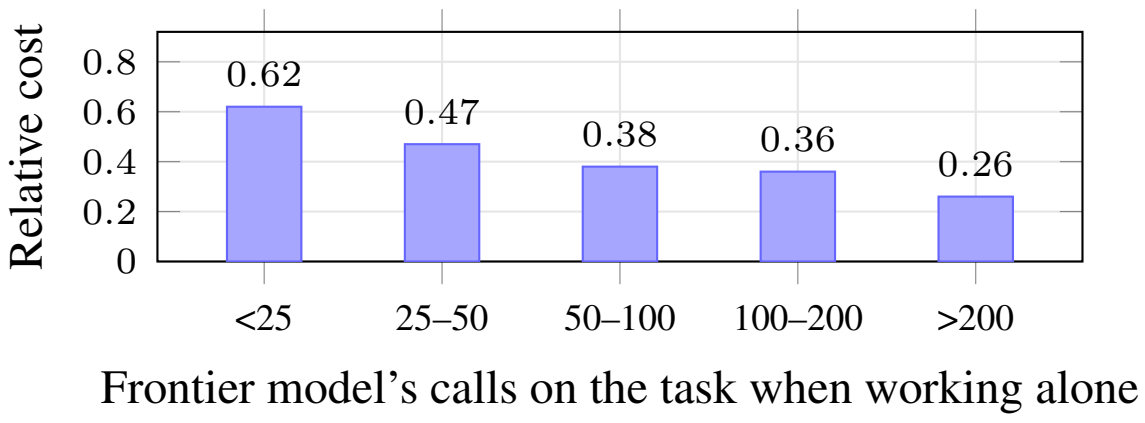

\paragraph{How do the savings vary across tasks?}
The saving on a task depends on how much work the frontier model would do on it alone. \Cref{fig:legwork} groups the
coding tasks by the number of calls that the frontier model makes without delegation. On tasks that need fewer than 25
calls, delegation brings frontier cost to about 60\% of working alone, and on 11\% of these tasks, it even costs more.
The relative cost keeps falling as tasks grow longer, to about a quarter for tasks that need more than 200 calls, where
delegation never costs more.  On search, the corresponding quantity is how much the frontier model reads (\Cref{sec:results}). The
saving can thus be anticipated from how long the frontier model works on a task alone.

\paragraph{Where do the savings come from?}
Claude Code already delegates by default, but to the frontier model itself. Without our harness, \solfull\ starts
subagents on 98\% of DeepSWE tasks, and these subagents, which also run on \solfull, make more calls than the main loop
and account for 53\% of its cost. \sysname\ moves this work to the SLM, which explains 80\% of the saving. The remainder
comes from the frontier model itself, which now takes 46 to 48 LLM turns per task instead of 84. It reads and edits code
about as often as before, but its test runs drop from about eight per task to one or two, and it no longer makes nearly all of the 23
calls per task that it spent on maintaining a task list and messaging its subagents. Since only reports return, the tool output that reaches the frontier model also shrinks, by 11 to 12\% on DeepSWE and by 13 to 24\% on \mbox{SWE-bench} Pro.

\paragraph{What work does the SLM take over?}
The SLM takes over the legwork of finding and checking, while the frontier model handles the main coding tasks itself. On
coding, a subagent typically searches the repository for the code involved in an issue, reads the relevant files, and
runs the tests or a small script to reproduce a bug or confirm a fix. Per DeepSWE task, the subagents of \qwen\ issue
about 24 code searches, read about 30 files or parts of files, run the tests 12 times, and run 5 scripts of their own.
The files they write, about four per task, are nearly all such \mbox{throwaway} scripts outside the repository. \mini\ works
the same way with more steps, and in 9 of 50 episodes, it also edits source files. On search, a subagent issues its own
BM25 searches, reads the documents it finds, and returns an answer with quoted evidence, its sources, and a confidence,
and the harness checks each quote against its source. In both cases, the frontier model receives findings that it can
verify cheaply, and it decides the answer on its own.

\begin{table}[t]
\centering
\small
\setlength{\tabcolsep}{3pt}
\begin{tabular}{@{}lccc@{}}
\toprule
\makecell[l]{Frontier\\+ subagents} & Solved & \makecell{LLM\\cost} & \makecell{LLM\\turns} \\
\midrule
\multicolumn{4}{@{}l}{\emph{Without our instruction}} \\
\quad \solsh\ alone & 29 & 100\% & \phantom{0}84 \\
\quad \solsh\ + \qwensh & 28 & \phantom{0}55\% & 101 \\
\quad \solsh\ + \minish & 22 & \phantom{0}51\% & \phantom{0}90 \\
\midrule
\multicolumn{4}{@{}l}{\emph{With our instruction}} \\
\quad \solsh\ + \solsh & 34 & \phantom{0}60\% & \phantom{0}52 \\
\quad \solsh\ + \qwensh & 29 & \phantom{0}33\% & \phantom{0}46 \\
\quad \solsh\ + \minish & 30 & \phantom{0}34\% & \phantom{0}48 \\
\bottomrule
\end{tabular}
\caption{With and without our instruction, on 50 DeepSWE tasks. \solsh\ alone runs its subagents on
\solsh. LLM cost is what \solsh\ is charged, including any \solsh\ subagents, relative to \solsh\ alone; it
excludes the SLM.}
\label{tab:anatomy}
\end{table}

\paragraph{Is the instruction necessary?}
\Cref{tab:anatomy} varies the instruction and the model that runs the subagents, and shows that the instruction is necessary, particularly for the weaker SLM.
Without the instruction, delegating to an SLM about halves frontier cost. With \qwen, the frontier model resolves 28 of
the 50 tasks at 55\% of the cost, but with \mini, it resolves only 22, seven fewer than on its own. With the
instruction, both SLMs match the frontier model alone at a third of the cost, so task performance no longer depends on
the SLM. With subagents still running on \solfull, the instruction alone brings frontier cost to 60\% with no loss in
task performance, and replacing these subagents with an SLM removes a further 27 points. Roughly three fifths of the
saving can thus be attributed to the instruction and two fifths to the SLM.

\paragraph{Does the strength of the SLM matter?}
Not much in our experiments, even though the two SLMs are far apart on their own: on the same 30 DeepSWE tasks, \qwen\ resolves 5
and \mini\ none. As subagents, however, they lead to the same task performance and nearly the same frontier cost
(\Cref{tab:coding}). Since the frontier model hands out narrow subtasks and handles the main coding tasks itself,
a weaker SLM mostly costs more SLM calls, 106 instead of 82 per DeepSWE task, rather than accuracy.

\begin{table}[t]
\centering
\small
\setlength{\tabcolsep}{2.5pt}
\begin{tabular}{@{}lcc@{}}
\toprule
Per DeepSWE task & \solsh & \opussh \\
\midrule
How it works alone & \makecell{long loop of\\edit, test, fix} & \makecell{plan first,\\then act} \\
Solved alone, of 50 & 29 & 34 \\
LLM turns, alone & 84 & 18 \\
\midrule
\multicolumn{3}{@{}l}{\emph{With an SLM, under our instruction}} \\
\quad Solved, of 50 & 29.5 & 38 \\
\quad LLM turns & 47 & 13 \\
\quad Turns handed to the SLM & 94 & 21 \\
\quad SLM turns before 1st LLM edit & 25\% & 59\% \\
\quad Words in each brief & 45 & 130 \\
\quad Frontier cost vs.\ LLM alone & 34\% & 69\% \\
\bottomrule
\end{tabular}
\caption{Two frontier models in \sysname\ on DeepSWE. With an SLM, values average the two SLMs. A brief is what the
frontier model writes for a subagent.}
\label{tab:twouses}
\end{table}

\paragraph{What changes with a different frontier model?}
The saving depends on how the frontier model works, as \Cref{tab:twouses} shows for \solfull\ and \opus. On its own,
\solfull\ works in a long loop of editing, testing, and fixing. It takes 84 LLM turns per DeepSWE task and makes its
first edit at turn 18. \opus\ instead plans before it acts, taking 18 turns and making its first edit at turn 7.
Delegation follows these habits. \solfull\ hands off much of its loop, mostly after it has started editing, so about
94 turns per task move to the SLM and its cost falls to 34\%. \opus\ delegates mainly before its first edit, to learn
about the code it is about to change, and writes longer briefs, about 130 words instead of 45. Only about 21 turns move
to the SLM, and its cost falls to 69\%, roughly in proportion to its own turns, which fall from 18 to 13. With either
SLM, \opus\ also resolves slightly more tasks than on its own. Thus, \sysname\ saves the most for a frontier model
that does much of its own legwork, while a frontier model that plans first uses the SLM mainly to inform its plan.

% ===========================================================================
\section{Related Work}
\label{sec:related}

Cascades such as FrugalGPT~\citep{chen2023frugalgpt} and routers such as RouteLLM~\citep{ong2024routellm} decide which
model answers; in \sysname, the frontier model always answers. In Minions~\citep{narayan2025minions}, a cloud model decomposes long-document tasks for a local model; we study
long-horizon agentic tasks in which the frontier model decides what to hand off. Orchestrator-worker designs are common
in agents,\footnote{\url{https://www.anthropic.com/research/building-effective-agents}} and \citet{belcak2025slm} argue that SLMs suit much of agentic work.

Concurrent with our work, \citet{samway2026teams} study teams of coding agents in which a lead agent
delegates subtasks to workers, and find that cheap models pay off as planners but not as implementers.\footnote{See also \url{https://cohere.com/blog/lessons-from-scaling-teams-of-frontier-coding-agents}} Their workers
are API models, and the team spends more compute for higher reward, while \sysname\ delegates to a local SLM to lower
frontier cost. Our ablation agrees, since without our instruction, the 2B SLM loses tasks.

% ===========================================================================
\section{Conclusion}

On \sysname, a frontier model spends a third as much on DeepSWE and 41\% as much on agentic search, with no loss in
task performance, and the saving grows with the legwork the frontier model would otherwise do. Claude Code
already delegates much of this legwork, but to the frontier model. A short instruction that leaves the main coding tasks
to the frontier model makes this work, and with it, even a 2B SLM loses no tasks. How much is saved also depends on the
frontier model. One that iterates at length saves the most, while one that plans first uses the SLM mainly to inform its
plan. These results support our thesis that a local SLM can take over much of the work of a frontier model, cutting its
cost without hurting task performance and without training.

% ===========================================================================
\section*{Limitations}
\label{sec:limitations}

\paragraph{Reliance on Claude Code.}
For coding, \sysname\ reuses the agent loop and subagents of Claude Code as they are, and adds only the harness around them: routing each model call by role to a frontier model or a local SLM, cache-aware serving on both
sides, and the instruction that steers delegation. For search, it also adds the delegation tool and the evidence check, and its SLM subagent uses the prompts and
tools of AREX-Turbo, whose final report cites a source for every claim, which is what the evidence
check needs.
The amount of harness engineering is thus modest, and our finding is that it suffices to remove most of the frontier cost.

\paragraph{Cost of the SLM.}
The SLM is not free. We report frontier cost because it is the metered, scarce resource, and both of our SLMs fit on hardware that a
user can own. At the API price of the same FP8 model on Chutes, \$0.039 per million input tokens,
a tenth of that for input repeated from the subagent's previous request, which is 86 to 90\% of its input in our runs, and \$1.25 per million output tokens, \qwen\
adds 3 to 10 cents per coding task, and the total stays at 38 to 57\% of the frontier model alone. No API serves \mini. We price the SLM for coding only, since
our search logs keep no per-call token counts.

\paragraph{Coverage of tasks and models.}
We study agentic coding and agentic search only, though these are two of the most important capabilities of today's
models. Our main results use \solfull\ at high reasoning effort. We test \opus\ as the frontier model on DeepSWE only, and the 2B
SLM on coding only. For coding, \sysname\ builds on Claude Code, which \opus\ is likely tuned for, so it may favor
\opus\ there. On agentic search, the frontier model uses each corpus's standard setup, and \solfull\ alone answers
193 of 200 BrowseComp-Plus questions, so it is not held back there. How much work is delegated depends on how well the
frontier model follows the instruction, which may differ for other frontier models.

\paragraph{Latency.}
Delegation makes tasks slower, about twice as slow at the median on short coding tasks, which may not suit interactive
use. How much slower depends on the hardware that serves the SLM. Our latencies were measured on shared GPUs of several
kinds and will vary with the GPU, the serving stack, and the load.

\paragraph{No claim of improved performance.}
Agent runs are not deterministic, and different frontier models use delegation differently. \solfull\ hands off
its legwork to save cost, while \opus\ uses its subagents to check its own work. A gain in task performance is
therefore not guaranteed for every frontier model or task set. We claim only that delegation causes no loss in task
performance, even where the delegating system solves more tasks than the frontier model alone.
The reduction in frontier cost, on the other hand, is consistent across task sets and models.

\clearpage
\bibliography{custom}

@article{chen2023frugalgpt,
  title   = {{FrugalGPT}: How to Use Large Language Models While Reducing Cost and Improving Performance},
  author  = {Chen, Lingjiao and Zaharia, Matei and Zou, James},
  journal = {Transactions on Machine Learning Research},
  year    = {2024},
  note    = {{arXiv}:2305.05176}
}

@inproceedings{madaan2023automix,
  title     = {{AutoMix}: Automatically Mixing Language Models},
  author    = {Aggarwal, Pranjal and Madaan, Aman and Anand, Ankit and Potharaju, Srividya Pranavi and Mishra, Swaroop and Zhou, Pei and Gupta, Aditya and Rajagopal, Dheeraj and Kappaganthu, Karthik and Yang, Yiming and Upadhyay, Shyam and Faruqui, Manaal and Mausam},
  booktitle = {Advances in Neural Information Processing Systems},
  year      = {2024},
  note      = {{arXiv}:2310.12963}
}

@inproceedings{yue2024cascades,
  title     = {Large Language Model Cascades with Mixture of Thought Representations for Cost-Efficient Reasoning},
  author    = {Yue, Murong and Zhao, Jie and Zhang, Min and Du, Liang and Yao, Ziyu},
  booktitle = {International Conference on Learning Representations},
  year      = {2024},
  note      = {{arXiv}:2310.03094}
}

@inproceedings{ding2024hybridllm,
  title     = {Hybrid {LLM}: Cost-Efficient and Quality-Aware Query Routing},
  author    = {Ding, Dujian and Mallick, Ankur and Wang, Chi and Sim, Robert and Mukherjee, Subhabrata and R{\"u}hle, Victor and Lakshmanan, Laks V. S. and Awadallah, Ahmed Hassan},
  booktitle = {International Conference on Learning Representations},
  year      = {2024},
  note      = {{arXiv}:2404.14618}
}

@inproceedings{ong2024routellm,
  title     = {{RouteLLM}: Learning to Route {LLMs} from Preference Data},
  author    = {Ong, Isaac and Almahairi, Amjad and Wu, Vincent and Chiang, Wei-Lin and Wu, Tianhao and Gonzalez, Joseph E. and Kadous, M. Waleed and Stoica, Ion},
  booktitle = {International Conference on Learning Representations},
  year      = {2025},
  note      = {{arXiv}:2406.18665}
}

@inproceedings{narayan2025minions,
  title     = {Cost-efficient Collaboration between On-device and Cloud Language Models},
  author    = {Narayan, Avanika and Biderman, Dan and Eyuboglu, Sabri and May, Avner and Linderman, Scott and Zou, James and R{\'e}, Christopher},
  booktitle = {Proceedings of the 42nd International Conference on Machine Learning},
  series    = {PMLR},
  volume    = {267},
  pages     = {45682--45719},
  year      = {2025},
  note      = {{arXiv}:2502.15964}
}

@article{belcak2025slm,
  title   = {Small Language Models are the Future of Agentic {AI}},
  author  = {Belcak, Peter and Heinrich, Greg and Diao, Shizhe and Fu, Yonggan and Dong, Xin and Muralidharan, Saurav and Lin, Yingyan Celine and Molchanov, Pavlo},
  journal = {arXiv preprint arXiv:2506.02153},
  year    = {2025}
}

@inproceedings{jimenez2024swebench,
  title     = {{SWE}-bench: Can Language Models Resolve Real-World {GitHub} Issues?},
  author    = {Jimenez, Carlos E. and Yang, John and Wettig, Alexander and Yao, Shunyu and Pei, Kexin and Press, Ofir and Narasimhan, Karthik},
  booktitle = {International Conference on Learning Representations},
  year      = {2024}
}

@article{deng2025swebenchpro,
  title   = {{SWE-Bench Pro}: Can {AI} Agents Solve Long-Horizon Software Engineering Tasks?},
  author  = {Deng, Xiang and Da, Jeff and Pan, Edwin and others},
  journal = {arXiv preprint arXiv:2509.16941},
  year    = {2025}
}

@article{deepswe2025,
  title   = {{DeepSWE}: Measuring Frontier Coding Agents on Original, Long-Horizon Engineering Tasks},
  author  = {Huang, Wenqi and Lee, Charley and Tng, Leonard and Ge, Serena},
  journal = {arXiv preprint arXiv:2607.07946},
  year    = {2026}
}

@article{chen2025browsecompplus,
  title   = {{BrowseComp-Plus}: A More Fair and Transparent Evaluation Benchmark of Deep-Research Agent},
  author  = {Chen, Zijian and Ma, Xueguang and Zhuang, Shengyao and Nie, Ping and Zou, Kai and Liu, Andrew and Green, Joshua and Patel, Kshama and Meng, Ruoxi and Su, Mingyi and Sharifymoghaddam, Sahel and Li, Yanxi and Hong, Haoran and Shi, Xinyu and Liu, Xuye and Thakur, Nandan and Zhang, Crystina and Gao, Luyu and Chen, Wenhu and Lin, Jimmy},
  journal = {arXiv preprint arXiv:2508.06600},
  year    = {2025}
}

@inproceedings{yang2018hotpotqa,
  title     = {{HotpotQA}: A Dataset for Diverse, Explainable Multi-hop Question Answering},
  author    = {Yang, Zhilin and Qi, Peng and Zhang, Saizheng and Bengio, Yoshua and Cohen, William W. and Salakhutdinov, Ruslan and Manning, Christopher D.},
  booktitle = {Proceedings of the 2018 Conference on Empirical Methods in Natural Language Processing},
  pages     = {2369--2380},
  year      = {2018}
}

@inproceedings{diao2025climb,
  title     = {{Nemotron-CLIMB}: Clustering-based Iterative Data Mixture Bootstrapping for Language Model Pre-training},
  author    = {Diao, Shizhe and Yang, Yu and Fu, Yonggan and Dong, Xin and Su, Dan and Kliegl, Markus and Chen, Zijia and Belcak, Peter and Suhara, Yoshi and Yin, Hongxu and Patwary, Mostofa and Lin, Yingyan Celine and Kautz, Jan and Molchanov, Pavlo},
  booktitle = {Advances in Neural Information Processing Systems},
  year      = {2025},
  note      = {{arXiv}:2504.13161}
}

@article{robertson2009bm25,
  title   = {The Probabilistic Relevance Framework: {BM25} and Beyond},
  author  = {Robertson, Stephen and Zaragoza, Hugo},
  journal = {Foundations and Trends in Information Retrieval},
  volume  = {3},
  number  = {4},
  pages   = {333--389},
  year    = {2009}
}

@inproceedings{lin2021pyserini,
  title     = {Pyserini: A {Python} Toolkit for Reproducible Information Retrieval Research with Sparse and Dense Representations},
  author    = {Lin, Jimmy and Ma, Xueguang and Lin, Sheng-Chieh and Yang, Jheng-Hong and Pradeep, Ronak and Nogueira, Rodrigo},
  booktitle = {Proceedings of the 44th International ACM SIGIR Conference on Research and Development in Information Retrieval},
  pages     = {2356--2362},
  year      = {2021}
}

@inproceedings{kwon2023vllm,
  title     = {Efficient Memory Management for Large Language Model Serving with {PagedAttention}},
  author    = {Kwon, Woosuk and Li, Zhuohan and Zhuang, Siyuan and Sheng, Ying and Zheng, Lianmin and Yu, Cody Hao and Gonzalez, Joseph E. and Zhang, Hao and Stoica, Ion},
  booktitle = {Proceedings of the 29th Symposium on Operating Systems Principles},
  year      = {2023}
}

@inproceedings{hu2024minicpm,
  title     = {{MiniCPM}: Unveiling the Potential of Small Language Models with Scalable Training Strategies},
  author    = {Hu, Shengding and Tu, Yuge and Han, Xu and others},
  booktitle = {Conference on Language Modeling},
  year      = {2024},
  note      = {{arXiv}:2404.06395}
}

@article{minicpm2025,
  title   = {{MiniCPM4}: Ultra-Efficient {LLMs} on End Devices},
  author  = {{MiniCPM Team}},
  journal = {arXiv preprint arXiv:2506.07900},
  year    = {2025}
}

@article{gupta2026deepsearchqa,
  title   = {{DeepSearchQA}: Bridging the Comprehensiveness Gap for Deep Research Agents},
  author  = {Gupta, Nikita and Chatterjee, Riju and Haas, Lukas and Tao, Connie and Wang, Andrew and Liu, Chang and Oiwa, Hidekazu and Gribovskaya, Elena and Ackermann, Jan and Blitzer, John and Goldshtein, Sasha and Das, Dipanjan},
  journal = {arXiv preprint arXiv:2601.20975},
  year    = {2026}
}

@article{rafiee2026trqa,
  title   = {{Total Recall QA}: A Verifiable Evaluation Suite for Deep Research Agents},
  author  = {Rafiee, Mahta and Soudani, Heydar and Abbasiantaeb, Zahra and Aliannejadi, Mohammad and Hasibi, Faegheh and Zamani, Hamed},
  journal = {arXiv preprint arXiv:2603.18516},
  year    = {2026}
}

@article{cmass2026,
  title   = {{CMASS}: A {ClimbMix} Agentic Search Suite for Evaluating Search Agents Across Heterogeneous Tasks},
  author  = {Sharifymoghaddam, Sahel and Gu, Lingwei and Ge, Yijun and Lin, Jimmy},
  journal = {arXiv preprint arXiv:2608.20317},
  year    = {2026}
}

@article{samway2026teams,
  title   = {Scaling Inference-Time Compute with Teams of Coding Agents},
  author  = {Samway, Keenan and D'souza, Daniel and Domingues, Omar Darwiche and Fadaee, Marzieh and Yi-Chern, Tan and Dziri, Nouha},
  journal = {arXiv preprint arXiv:2610.17471},
  year    = {2026}
}

@article{baai2026arex,
  title   = {{AREX}: Towards a Recursively Self-Improving Agent for Deep Research},
  author  = {Lu, Shuqi and Li, Chaofan and Luo, Kun and Zhang, Zhang and Wang, Hui and Xiao, Hongwang and Xiong, Lei and Wang, Jiahao and Wang, Sen and Jiang, Xiyan and Li, Wanli and Hu, Yuyang and Qian, Hongjin and Yan, Bingyu and Chen, Jianlyu and Xia, Ziyi and Shao, Yingxia and Liu, Kang and Dou, Zhicheng and He, Di and Li, Chaozhuo and Ye, Qiwei and Wang, Zhongyuan and Liu, Zheng},
  journal = {arXiv preprint arXiv:2607.21461},
  year    = {2026}
}

% ===========================================================================
\clearpage
\appendix

\section{Prompts}
\label{app:prompts}

\paragraph{Coding instruction.} We supply this instruction to Claude Code as a user-level \texttt{CLAUDE.md} file.
\begin{promptbox}
\# Working with subagents

Subagents (the Agent tool) run on a small, fast model and cost far less than you. Every step you take re-reads the whole
conversation, so the cheapest plan is the one with the fewest steps of your own.
\pitem{Delegate investigation (find the code involved in the issue and explain how it works; write a minimal script
that reproduces the bug and report its output) and verification (run the reproduction and the relevant tests; report
only what fails). Make the code changes yourself.}
\pitem{Run subagents in the foreground (do not use run\_in\_background) and launch independent ones in the same turn,
so their reports come back together and nothing needs polling.}
\pitem{Ask each subagent for a short report, at most 150 words: only the facts you need, with file paths, line numbers
and exact error messages.}
\pitem{Do not keep a task list (TaskCreate / TaskUpdate / TaskList / TaskGet) for this job; keep your plan in your own
messages.}
\pitem{Subagent reports can be wrong: check a fact yourself before your fix depends on it.}
\end{promptbox}

\paragraph{Search instruction.} We append this instruction to the system prompt of the frontier model.
\begin{promptbox}
You have a research assistant: the Delegate tool hands it a focused subtask; it searches the same corpus with its own
search engine and returns a short report with evidence, the docids and its confidence. It is fast, and its
high-confidence reports are usually right. Work through it:
\pitem{Delegate all searching, including checks of a candidate ("does the corpus confirm that X ...?"). Run independent
subtasks in parallel.}
\pitem{Do not run broad searches or read whole documents yourself. To confirm one fact, use a single targeted search, or
read only the part of the document around it (get\_document with an offset).}
\pitem{Each report ends with an automatic lookup of its evidence in the cited documents: a located passage is verbatim
there.}
\pitem{Give your final answer as soon as the evidence covers every condition of the question.}
\end{promptbox}

\paragraph{Example briefs.} Below is the first brief that each frontier model wrote for a \qwen\ subagent on the same DeepSWE
task, which adds module loading flags to the interpreter of the ABS language. The brief of \solfull:
\begin{promptbox}
Inspect ABS module require loader, cache, environments, tests and project instructions. Read-only investigation; do
not edit. Identify files/functions to change and relevant test commands, including how environment stderr and runtime
values work. Short report at most 150 words with paths and line numbers.
\end{promptbox}
\noindent The brief of \opus, abridged:
\begin{promptbox}
In the Go repo at /app (the ABS programming language interpreter), report facts only (max 200 words, with file paths
and line numbers; do not modify files):
\pitem{How the builtin require(path) is implemented today: the function name/file, how it resolves the path (relative
to what directory? how the current file's directory is tracked [...]), whether there's a module cache (variable name,
key type), and how the module file is lexed/parsed/evaluated (which helper it calls).}
\pitem{How builtin functions are registered (the map name/file, function signature, how they access the environment
[...]).}
\pitem{The CLI entrypoint: main.go and repl/repl.go BeginRepl(args []string, version string) [...]}
\end{promptbox}
\noindent The later testing brief of \opus\ asks the subagent, for every failing test outside the new test files, to check whether it also
fails on the base branch and to report whether the failure is pre-existing.

\section{Run Settings}
\label{app:settings}

\begin{table}[h]
\centering
\small
\begin{tabular}{@{}lp{0.52\columnwidth}@{}}
\toprule
Setting & Value \\
\midrule
Claude Code version & 2.1.232 \\
Tools & default, except web search and web fetch \\
Frontier model context & up to 1M tokens \\
SLM context & 131k tokens \\
Turn limit & 100 turns \\
Tool time limit & 60 minutes (90 on DeepSWE) \\
\bottomrule
\end{tabular}
\caption{Settings shared by all coding runs.}
\label{tab:settings}
\end{table}

\paragraph{Coding.}
\Cref{tab:settings} lists the settings that all coding runs share. Both SLMs think before acting on coding, while on
search, \qwen\ answers without thinking.

\paragraph{Search subagent.}
On search, the SLM subagent works in our own loop of at most 80 steps with four tools from AREX-Turbo:
\begin{itemize}[leftmargin=*,itemsep=1pt,topsep=2pt,parsep=0pt,partopsep=0pt]
  \item \texttt{search} takes up to five queries and returns the top 10 BM25 hits for each with their best-matching
        passage, through the ClimbMix search API or, on BrowseComp-Plus, a BM25 index of the corpus files with the
        benchmark's default parameters.
  \item \texttt{visit} returns the passages of a document that best match a stated goal.
  \item \texttt{update\_context} replaces the history of the subagent with its own notes.
  \item \texttt{finish} returns the answer, its evidence with sources, and a confidence.
\end{itemize}
We took these prompts and tools as published, leaving out the Google Scholar tool of AREX-Turbo, and did not tune them.

\paragraph{Frontier model on search.}
The frontier model runs on the BrowseComp-Plus corpus with the system prompt and the Bash and Read tools of Claude
Code, taken from a recorded Claude Code session, in our own loop, which adds the \texttt{Delegate} tool. On ClimbMix, it
uses the \texttt{search} and \texttt{get\_document} tools of the official BrowseComp-Plus agent. With delegation, the
output of the frontier model's own tool calls on BrowseComp-Plus is cut at 12,000 characters instead of Claude Code's
30,000. The cut applied to 18 of its 395 own calls.

\paragraph{Hardware.}
Each SLM replica serves several concurrent episodes, almost always from a single GPU. For coding,
\qwen\ ran on NVIDIA H200 and RTX PRO 6000 Blackwell GPUs, and \mini\ on H200, RTX PRO 6000 Blackwell, RTX 6000 Ada, and
L40S GPUs. The \opus\ runs also used H100 and RTX A6000 GPUs, with one \qwen\ replica split across two A6000s. For
search, \qwen\ ran mostly on RTX PRO 6000 and H100 GPUs. Due to budget constraints, we served \opus\ through Claude
subagents, each call with a clean context.

\paragraph{Choosing the tools.}
In our experiments, the tools were set by the task set. Afterwards, we asked \solfull\ at low effort to choose between
the two tool sets with the following prompt:
\begin{promptbox}
Here is a task. <task> ... </task> Do not do the task. Decide which kind of task it is and answer with exactly one
word: coding if it asks you to change code in a software repository, or search if it asks you to find the answer to a
question.
\end{promptbox}
\noindent For all 869 tasks and questions, it chose the tools that our experiments used. The question costs about
\$0.0005 per task, under one point of relative cost on every task set, which our cost figures leave out.

\paragraph{Reruns and later runs.}
Claude Code compacts a long conversation to keep its context short. An episode of the frontier model that compacted
and then failed was rerun with its full 1M-token context, in every system alike. Search questions that failed for
infrastructure reasons (8 of 694) were rerun with identical settings, and no question was rerun because of its answer.
Runs added after our main experiments, namely the reruns after compaction and the runs of \Cref{sec:analysis}, served
both SLMs on a shared vLLM cluster with the same weights, behind a proxy that applies the same request conversion as
our own servers.

\section{Task Sampling}
\label{app:sampling}

Coding tasks were sampled with Python's \texttt{random.Random(20261020)} from the tasks that all compared systems
finished without an infrastructure failure, namely 50 SWE-bench Verified tasks (all such tasks among those not used to draft
the instruction), 75 SWE-bench Pro tasks, and 50 DeepSWE v1.1 tasks. The comparison of the two SLMs on their own uses 30
DeepSWE tasks sampled with the same seed from those that both finished without an infrastructure failure. The 200
BrowseComp-Plus questions were sampled with seed 20261019 from held-out questions. The sampled task lists are released
with the paper.

\end{document}